\pdfoutput=1

\documentclass[11pt]{article}

 \usepackage{emnlp2021}

\usepackage{times}
\usepackage{latexsym}

\usepackage[T1]{fontenc}

\usepackage[utf8]{inputenc}

\usepackage{graphicx}
\usepackage{microtype}
\usepackage{changepage}

\usepackage[utf8]{inputenc}
\usepackage[arabic,USenglish]{babel}
\title{TunBERT: Pretrained Contextualized Text Representation for Tunisian Dialect}

\author{Abir Messaoudi \\ \bf{Hatem Haddad} \\ \bf{Moez BenHajhmida} \\ \bf{Malek Naski}\\
  \it{iCompass} \\
  \And
  
  Ahmed Cheikhrouhou \\ \bf{Nourchene Ferchichi}\\ \bf{Abir Korched}   \\   \bf{Faten Ghriss}    \\\bf{Amine Kerkeni} \\

  \it{InstaDeep}  \\
  
}

\begin{document}
\maketitle
\begin{abstract}
Pretrained contextualized text representation models learn an effective representation of a natural language to make it machine understandable. After the breakthrough of the attention mechanism, a new generation of pretrained models have been proposed achieving good performances since the introduction of the Transformer. Bidirectional Encoder Representations from Transformers (BERT) has become the state-of-the-art model for language understanding.  Despite their success, most of the available models have been trained on Indo-European languages  however similar research for under-represented languages and dialects remains sparse.
 In this paper, we investigate the feasibility of training monolingual Transformer-based language models for under represented languages, with a specific focus on the Tunisian dialect. We evaluate our language model on sentiment analysis task, dialect identification task and reading comprehension question-answering task. We show that the use of noisy web crawled data instead of structured data (Wikipedia, articles, etc.) is more convenient for such non-standardized language. Moreover, results indicate that a relatively small web crawled dataset leads to performances that are as good as those obtained using larger datasets. Finally, our best performing  TunBERT model reaches or improves the state-of-the-art in all three downstream tasks. We release the TunBERT pretrained model  and the datasets used for fine-tuning\footnote{To preserve anonymity, a link to Github repository will be added to the camera-ready version if the paper is accepted.}.
\end{abstract}

\section{Introduction}
In the last decade, natural language understanding has gained interest owing to the available hardware and data resources and to the evolution of the pretrained contextualized text representation models. These models learn an effective representation of a natural language to make it machine understandable. Word2Vec \cite{Word2vec} has been one of the first proposed approaches where words were represented according to their  semantic property. Next, ELMO \cite{elmo} combined the previous model with BiLSTM in order to deal with the polysemy problem. Afterwards, the pretraining models have been firstly proposed with ULMFit \cite{ulmfit} where they were fine-tuned for downstream tasks. These models have achieved good performances but they did not support long-term and multiple contexts of the words.\\

After the breakthrough of the attention mechanism \cite{attention}, a new generation of pretrained models have appeared. They have achieved tremendous performances since the introduction of the Transformers \cite{gpt1}. Besides, the Bidirectional Encoder Representations from Transformers (BERT) \cite{BERT} has been  unleashed to become the state-of-the-art model for language understanding and gave new inspiration to further development in the Natural Language Processing (NLP) field. Accordingly, most languages have their own BERT-based language models. Specifically, the Arabic language has multiple language models: AraBERT \cite{ARABERT}, GigaBERT \cite{lan2020gigabert}, and multilingual cased BERT model (hereafter mBERT) \cite{pires-etal-2019-multilingual} which was simultaneously pretrained on 104 languages. 

Arabic language has more than 300 million native speakers around the world and it's used as a native language in 26 countries. The official form of Arabic is called Modern Standard Arabic (MSA). Although, each country has one or more locally Arabic spoken language, called Dialect. The people in Tunisia use the Tunisian dialect \cite{Chayma2020} in their daily communications, most of their media (TV, radio, songs, etc), and on the internet (social media, forums). Yet, this dialect is not standardized which means there is no unique way for writing and speaking. Added to that, it has its proprietary lexicon, phonetics, and morphological structure as shown in Table \ref{tunisianexamples}. \\

\begin{table*}
\centering
\begin{tabular}{lll}
\hline
\textbf{Tunisian} &  \textbf{MSA} & \textbf{English}  \\
\hline
\textRL{\foreignlanguage{arabic}{محلاها هالغناية}} &\textRL{\foreignlanguage{arabic} { ما أحلى هذه الأغنية}}  &  How nice is this song \\
\textRL{\foreignlanguage{arabic}{ماتعجبنيش كيفاش تتصرف}} & \textRL{\foreignlanguage{arabic}{ لا تعجبني تصرفاتها}} &I don't like how she behaves \\
\textRL{\foreignlanguage{arabic}{  وقتاه يبدا الماتش}} & \textRL{\foreignlanguage{arabic}{  متى تبدأ المباراة}} &When does the match start \\ 

\hline
\end{tabular}
\caption{Examples of Tunisian sentences with their MSA and English translation.}\label{tunisianexamples}
\end{table*}
The need for a robust language model for Tunisian dialect has become crucial to develop natural-language-processing-based applications (translation, information retrieval, sentiment analysis, etc). To the best of our knowledge, there is no such model proposed yet in literatures.\\





In this paper, we describe the process of pretraining a Pytorch implementation of NVIDIA BERT language model\footnote{\url{https://github.com/NVIDIA/DeepLearningExamples/tree/master/PyTorch/LanguageModeling/BERT}}, called TunBERT (Tunisian BERT), trained on only 67.2 MB web-scraped dataset. We systematically compare our pre-trained model on three NLP downstream tasks;
that are different in nature: (i) Sentiment Analysis (SA),
(ii) Tunisian dialect identification (TDI), and (iii) Reading Comprehension Question-Answering (RCQA);  against mBERT \cite{BERT}, AraBERT \cite{ARABERT}, GigaBERT \cite{lan2020gigabert} and the state of the art performances when available.
Our contributions can be summarized as follows: 
\begin{itemize}
\item First release of a pretrained BERT
model for the Tunisian dialect using a Tunisian large-scale web-scraped dataset.
\item TunBERT application to three NLP downstream
tasks: Sentiment Analysis (SA), Tunisian dialect identification  (TDI) and  Reading Comprehension Question-Answering (RCQA).
\item Empirical evaluations illustrate that small and diverse Tunisian training dataset can achieve similar performance compared to several baselines including previous multilingual and single-language approaches trained on large-scale corpora.
\item  Publicly releasing TunBERT and the used datasets on popular NLP libraries\footnote{To preserve anonymity, a link to Github repository will be added to the camera-ready version if the paper is accepted.}. 
\end{itemize}

The rest of the paper is structured as follows. Section 2.
provides a concise literature review of previous work on monolingual and multilingual
language representation. Section 3. describes the used methodology to develop TunBERT. Section 4. describes the downstream tasks and
benchmark datasets that were used for evaluation. Section 5.
presents the experimental setup and discusses the results.
Finally, section 6. concludes and points to possible directions for future work.

\section{Related Works}
Contextualized word representations, such as BERT \cite{BERT}, RoBERTa \cite{RoBerta} and more
recently ALBERT \cite{ALBERT}, improved the representational power of word embeddings such as word2vec \cite{Word2vec}, GloVe \cite{Glove} and fastText \cite{fastext} by taking context into account. Following
their success, the large pretrained language models were extended to the multilingual setting such as  mBERT \cite{pires-etal-2019-multilingual}. In \cite{XLM}, authors showed that multilingual
models can obtain results competitive with monolingual
models by leveraging higher quality data
from other languages on specific downstream tasks. Nevertheless, these models have used large scale pretraining corpora and consequently need high computational cost. 

Recently, non-English monolingual models have
been released:  RobBERT for Dutch \cite{Robbert}, FlauBERT \cite{Flaubert} and CamemBERT for French  \cite{CAMEMBERT}, \cite{canete2020spanish} for Spanish and  \cite{finnish} for Finnish. In  \cite{CAMEMBERT}, authors showed that their French model trained on a
4 GB performed similarly to same model trained on the 138GB. They also concluded that a model trained on a  Common-
Crawl-based corpus performed consistently better than the one
trained on the French Wikipedia. They suggested that a  4 GB heterogeneous dataset
in terms of genre and style is large enough as a pretraining dataset  to reach
state-of-the-art results with the BASE architecure,
better than those obtained with mBERT (pretrained
on 60 GB of text). In \cite{finnish}, a Finnish BERT model trained from scratch outperformed mBERT for three reference tasks (part-of-speech tagging, named entity recognition, and dependency parsing). Authors suggested that a language-specific deep transfer learning
models for lower-resourced languages can outperform multilingual BERT models. 

Compared to the increasing studies of contextualized word representations in Indo-European languages, similar research for Arabic language is still very limited. AraBERT \cite{ARABERT}, a BERT-based model,  was released using a pre-training dataset of 70 million sentences, corresponding to 24 GB of text covering news from different Arab
media. AraBERT was pre-trained on a TPUv2-8
pod for 1,250K steps. It achieved state-of-the-art performances on three Arabic tasks: Sentiment Analysis, Named
Entity Recognition, and Question Answering. Nevertheless, the pre-trained dataset is mostly a MSA based. Authors concluded that there is a need for pretrained models that can tackle a variety of  Arabic dialects. Lately, GigaBERT  \cite{lan2020gigabert} customized bilingual language model for English and Arabic has outperformed AraBERT in several downstream tasks.

\section{TunBERT}
In this section, we describe the training Setup and pretraining data that was used for TunBERT.

\subsection{Training Setup}
TunBERT model is based on the  Pytorch implementation of NVIDIA NeMo BERT
\footnote{\url{https://github.com/NVIDIA/DeepLearningExamples/tree/master/PyTorch/LanguageModeling/BERT}}. The model was pre-trained using 4 NVIDIA Tesla V100 GPUs for 1280K steps. The pretrained model characteristics are shown in Table \ref{pretraining model}. Adam optimizer was used, with
a learning rate of 1e-4, a batch size of 128, a maximum sequence length of 128 and a masking probability of 15\%. Cosine annealing was used for learning rate scheduling with a warm-up ratio of 0.01.  Training took 122 hours and 25 minutes for 330 epochs over all the tokens. \\
The model was trained on two unsupervised prediction tasks using a large Tunisian text corpus: The Masked Language Modeling (MLM) task and the Next Sentence Prediction (NSP) task. \\
For the MLM task, 15\% of the words in each sequence are replaced with a [MASK] token. Then, the model attempts to predict the original masked token based on the context of the non-masked tokens in the sequence. For the NSP task, pairs of sentences are provided to the model. The model has to predict if the second sentence is the subsequent sentence in the original document. In this task, 50\% of the pair sentences are subsequent to each other in the original document. The remaining 50\% random sample sentences are chosen from the corpus to be added to the first sentence.

\subsection{Pre-training Dataset}
Because of the lack of available Tunisian dialect data (books, wikipedia, etc.), we use a  web-scraped dataset extracted from social media, blogs and websites consisting of 500k sentences of text, to pretrain the model.
The extracted data was  preprocessed by removing links, emoji and punctuation symbols. Then, a filter was applied to ensure that only Arabic scripts are included. Pretraining dataset statistics are presented in Table \ref{pretraining dataset}. The training dataset size is 67.2 MB.

\begin{table}
\centering
\begin{tabular}{lrl}
\hline
  \textbf{\#Uniq Words} & \textbf{\#Words} & \textbf{\#Sentences} \\
\hline
8,256K & 48,233K   &  500K \\

\hline
\end{tabular}
\caption{\label{pretraining dataset} Pretraining dataset statistics. }
\end{table}


\begin{table}
\centering
\begin{tabular}{lrl}
\hline
 \textbf{\#Layers} & \textbf{Hidden Size} & \textbf{\#self-attention heads} \\
\hline
 12 & 768  & 12   \\

\hline
\end{tabular}
\caption{\label{pretraining model} Pretrained model configuration. }
\end{table}

\section{Evaluation}
We measure the performance of TunBERT by evaluating it on three tasks: Sentiment Analysis, Dialect identification and Reading Comprehension Question-Answering. Fine-tuning was done independently using
the same configuration for all tasks. We do not run extensive
grid search for choosing the best hyper-parameters due to computational
and time constraints. We applied a configuration commonly used in the literature. We use the splits provided
by the datasets authors when available and the standard
80 \% and 20\% when not.

\subsection{Sentiment Analysis}
For the sentiment analysis task, we  used two manually annotated Tunisian Sentiment Analysis datastes: 
\begin{itemize}
    \item Tunisian Sentiment Analysis Corpus (TSAC) \cite{medhaffar-etal-2017-sentiment} obtained from
Facebook comments about popular TV shows. The TSAC dataset is composed of comments based on  Latin scripts, Arabic scripts and emoticons. We use only the Arabic script comments.
\item Tunisian Election Corpus (TEC) \cite{TEC} obtained from tweets about Tunisian elections in 2014. Beside Tunisian content, TEC dataset content is also composed of MSA content. 
\end{itemize}
 Statistics of the TSAC and TEC are shown in Table \ref{SAdataset}.


\begin{table}
\centering
\begin{tabular}{lrl}
\hline
\textbf{Dataset} &  \textbf{TSAC} & \textbf{TEC}  \\
\hline
\#Negative& 4175  & 1799 \\
\#Positive& 3277 &   1244  \\
\#Train &  4680   & 1947 \\
\#Dev &  1170 &  487 \\
\#Test & 1516  & 609  \\
\hline
\end{tabular}
\caption{\label{SAdataset} TSAC and TEC Sentiment analysis datasets statistics. }
\end{table}

\subsection{ Tunisian Dialect identification} 


This task focuses on identifying the Tunisian dialect of a given text from other Arabic dialects, especially on social media sources where there is no established standard orthography like MSA. First attempts to tackle the challenge identified 5 Arabic dialects categories in addition to MSA: Maghrebi, Egyptian, Levantine, Gulf, and Iraqi \cite{zaidan}. \cite{el-haj-etal-2018-arabic} proposed 4 Arabic dialects categories by merging the Iraqi with the Gulf. Tunisian dialect was classified  into the Maghrebi dialect along with the Algerian, Moroccan, and other dialects. Nevertheless, even if the Maghrebi vocabulary is
pretty much similar throughout North African countries, many differences exist not only at the phonetic level \cite{smaili} but also at the lexical, morphological and syntactic levels \cite{Horesh}. 

For evaluation, two sub-tasks were performed:
\begin{itemize}
    \item Identification of Tunisian dialect from other Arabic dialects (TADI): this is a binary classification task: Tunisian dialect and Non Tunisian dialect from an Arabic dialectical dataset. We used the Nuanced Arabic Dialect Identification (NADI) shared task dataset with a total of  21,000 tweets, covering 21 Arab countries. NADI is an imbalanced dataset in which the training includes only 747 Tunisian tweets and the remaining tweets cover other dialects. Consequently, this dataset is unbalanced. To solve this issue, we created a new dataset TADI (Tunisian and Arabic Dialect Identification) by including a sub-set of TSAC dataset as Tunisian comments to have the same number of tweets for the Tunisian dialect as same as the other dialects as shown in Table \ref{dialectstatistics}.

\begin{table}
\centering
\begin{tabular}{lrl}
\hline
\textbf{Dataset}  & \textbf{TADI} & \textbf{TAD} \\
\hline
\#Train & 40500   &  3200    \\
\#DEV &  2396  &  400   \\
\#Test &  7192  &  400   \\

\hline
\end{tabular}
\caption{\label{dialectstatistics} TADI and TAD dataset statistics. }
\end{table}
    \item Identification of Tunisian dialect and Algerian dialect (TAD dataset): for this sub-task we used the Multi-Arabic Dialect Applications and Resources (MADAR) dataset \cite{Bouamor-etal-2018} . More specifically, we used the shared task dataset to target a large set of dialect labels at country level. We filtered the dataset of dialect labels at country level \cite{MADAR} to only keep Tunisia and Algerian labeled data as shown in Table \ref{dialectstatistics}.

\end{itemize}
\subsection{Reading Comprehension Question-Answering}
Open-domain Question-Answering (QA) task has been intensively studied to evaluate the language Understanding performances of the models. This task takes as input a textual question to look for correspondent answers within a large textual corpus. In \cite{abs-1906-05394}, two MSA QA datasets has been proposed. However, to the best of our knowledge, no study was previously made for such a task for any Arabic dialect.\\

\begin{figure*}
  \includegraphics[width=\textwidth]{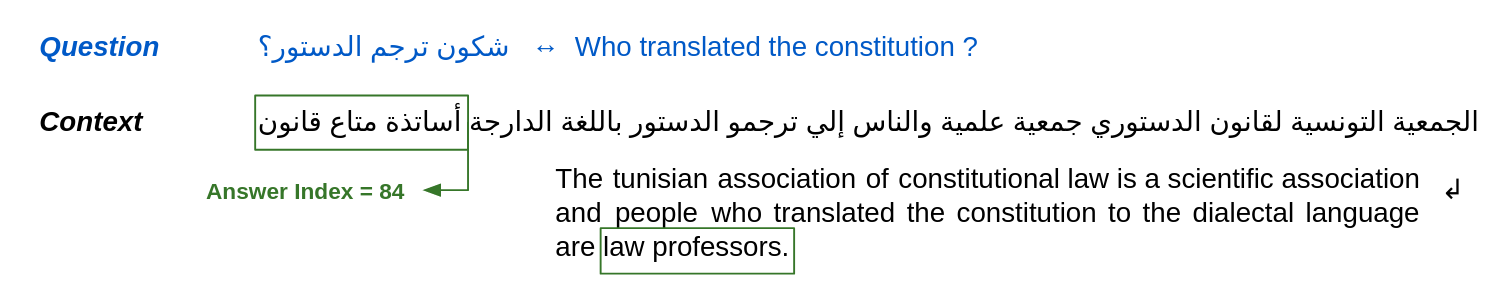}
  \caption{TRCD dataset example with its corresponding English translation. }
  \label{trcd-fig}
\end{figure*}

For this task, we built TRCD (Tunisian Reading Comprehension Dataset) as Question-Answering dataset for Tunisian dialect. We used a dialectal version of the Tunisian constitution following the guideline in \cite{Chen2017ReadingWT}. It is composed of 144 documents where each document has exactly 3 paragraphs and three Question-Answer pairs are assigned to each paragraph. Questions were formulated by four native speaker annotators and each question should be paired with a paragraph as shown in  Figure \ref{trcd-fig}). \\

To the best of our knowledge, this is the first Tunisian dialect dataset for the Question-Answering task. TRCD dataset statistics are showed in Table \ref{trcdstatistics}.

\begin{table}
\centering
\begin{tabular}{lrrl}
\hline
\textbf{Dataset}  & \textbf{\#Document} & \textbf{\#Paragraph} & \textbf{\#QA} \\
\hline
\#Train & 114   &  342 &  1026    \\
\#Dev &  15  &  45  &  135  \\
\#Test &  15  &  45 &  135   \\

\hline
\end{tabular}
\caption{\label{trcdstatistics} TRCD statistics. }
\end{table}

\section{Experiments and Discussion}

\subsection{Tunisian Sentiment Analysis}
 The efficiency of TunBERT language model was evaluated
against mBERT, AraBERT and GigaBERT language models and the state of the art performances when available. The
obtained performances of Tunisian Sentiment Analysis using TunBERT were further
compared against the baseline systems that tackled
the same datasets (word embeddings (word2vec), document embeddings (doc2vec) and Tw-StAR \cite{syntax}) and  listed in Table \ref{TSACresults} and Table \ref{TECresults}.

\begin{table}

\centering
\begin{adjustwidth}{-9mm}{}
\begin{tabular}{llrl}
\hline
 \textbf{Model} &    \textbf{Accuracy}&  \textbf{F1.macro
 } \\
\hline
\cite{medhaffar-etal-2017-sentiment} &   78\%  & 78\% \\
word2vec  \cite{syntax}& 77.4\%  & 78.2\% \\
doc2vec  \cite{syntax}& 57.2\%  & 61.7\% \\

  Tw-StAR \cite{syntax} & 86.5\% & 86.2\%   \\
 mBERT &  92.21\%  & 91.03\%  \\
  GigaBERT  & 94.92\% & 93.39\% \\
  AraBERT & 95.63\%   & 94.91\% \\
  TunBERT & \textbf{96.98\%}   & \textbf{96.98}\%\\

\hline
\end{tabular}
 \end{adjustwidth}
\caption{\label{TSACresults} TSAC results. }
\end{table}

\begin{table}
\centering
\begin{tabular}{llrl}
\hline
 \textbf{Model} &    \textbf{Accuracy}&  \textbf{F1.macro
 } \\
\hline
\cite{TEC} &  71.1\%  & 63\% \\
word2vec \cite{syntax}& 61.9 \% & 58.4\% \\
doc2vec \cite{syntax}& 62.2\%  & 56.4\% \\

  Tw-StAR \cite{syntax} & \textbf{88.2\%} &   \textbf{87.8\%} \\
 mBERT &  58.45\%  & 36.89\% \\
  GigaBERT  & 71.75\%	& 65.32\% \\
  AraBERT &  79.14\%	  &  72.57\%\\

  TunBERT & 81.2\%   & 76.45\%\\

\hline
\end{tabular}
\caption{\label{TECresults} TEC results. }
\end{table}

The results in Table \ref{TSACresults} illustrate the outperformance
of the pretrained contextualized text representation models over the previous techniques namely word2vec and doc2vec. TunBERT achieved the best performance on the TSAC dataset. It reached 92.98\% as F1.macro which is a high result comparing to to 78.2\%,
61.7\% and 86.2\% scored by word2vec, doc2vec and Tw-StAR, respectively. The results show that TunBERT also outperform pretrained language models: mBERT, GigaBERT and AraBERT.


Likewise,  Table \ref{TECresults} illustrates the outperformance of BERT-based LM against other techniques with the TEC dataset. Nevertheless, the best performances was achieved by Tw-StAR. For instance, the best achieved  Tw-StAR F1.macro was in TEC dataset with a value of 87.8\% compared to 76.45\%,
and 72.57\% scored by TunBERT  and AraBERT, respectively. This could be explained by the noisy nature of TEC dataset with a mixed Tunisian and MSA content. Results using mBERT achieved the worst performances could demonstrate that mBERT is not suitable for noisy data. The results showcase also the outperformance of TunBERT  over the other pretrained language models. 

\subsection{Tunisian Dialect identification}

For Tunisian Dialect identification, the
results in Table \ref{TADIcomparison} show that the  TunBERT language model outperform other state-of-the-art language models. Indeed, our model achieved a F1.macro of 87.14\%  compared to 68.93\% achieved by mBERT. TunBERT also outperforms the Arabic language pre-trained BERT AraBERT. Likewise, it has achieved a F1.macro of 93.25\% for the Tunisian-Algerian dialects identification task outperforming the other used language models as shown in Table \ref{TADcomparison}.

\begin{table}
\centering
\begin{tabular}{lrl}
\hline
 \textbf{Model} &    \textbf{Accuracy}&  \textbf{F1.macro
 } \\
\hline
 mBERT &  75,21\%& 68.93\% \\
  AraBERT &   79.57\% & 76.7\%   \\
  GigaBERT & 72.67\%	 & 	65.3\% \\

  TunBERT  &  \textbf{87.46\%}  & \textbf{87.14\%}  \\

\hline
\end{tabular}
\caption{\label{TADIcomparison} TADI results }
\end{table}

\begin{table}
\centering
\begin{tabular}{llrl}
\hline
 \textbf{Model} &    \textbf{Accuracy}&  \textbf{F1.macro
 } \\
\hline
 mBERT &  86.75\%& 86.4\% \\
  AraBERT &  87.5\%	 \% & 87.37\%   \\
  GigaBERT  &  & 0\%  \\

  TunBERT & \textbf{93.3\%}    & \textbf{93.25\%}  \\
\hline
\end{tabular}
\caption{\label{TADcomparison} TAD results }
\end{table}

\subsection{Reading Comprehension Question-Answering}

Fine-tuning TunBERT on the Tunisian Reading Comprehension Dataset did not give impressive results (Exact match of 2.17\%, F1 score of	13.66\% and a Recall of 	22.59\%). Comparable results were obtained for  GigaBERT (Exact match of 0.7\%, F1 score of	14.02\% and a Recall of	21.65\%). MBERT gave slightly better results (Exact match of 4.25\%, F1 score of	22.6\% and a Recall of31.3). Meanwhile, we noticed good results for AraBERT (Exact match of 26.24\%, F1 score of 58.74\% and a Recall of	63.96\%). \\

Adding a pre-training step on an MSA reading comprehension dataset (In our case the Arabic-SQuAD dataset \cite{abs-1906-05394}) made great improvements in all of the models performances, especially for the TunBERT.
The strategy was to use the pre-trained language model, fine-tune it for few epochs on the MSA dataset, then use the best checkpoint to train and test on the TRCD dataset. Following this stategy, TunBERT acheived great results with an Exact match of 27.65\%, an F1 score of 60.24\% and a Recall of 82.36\%, as shown in Table
\ref{TRCDcomparison}.

\begin{table*}
\centering
\begin{tabular}{|l|ccc|ccc|}
\hline
\multicolumn{1}{|c|}{\begin{tabular}[c]{@{}c@{}}Finetuning \\ datasets\end{tabular}} & \multicolumn{3}{c|}{TRCD dataset}                & \multicolumn{3}{c|}{\begin{tabular}[c]{@{}c@{}}Arabic SQuAD \\ and TRCD\end{tabular}} \\ \hline
Language Models                                                                                & Exact match    & F1 score       & Recall         & Exact match                 & F1 score                   & Recall                     \\ \hline
mBERT                                                                               & 4.25           & 22.6           & 31.3           & 29.07                       & 60.86                      & 62.18                      \\
AraBERT                                                                              & \textbf{26.24} & \textbf{58.74} & \textbf{63.96} & 24.11                       & 63.53                      & 70.43                      \\
GigaBERT                                                                             & 0.7            & 14.02          & 21.65          & 29.78                       & 62.44                      & 66.34                      \\
TunBERT                                                                      & 2.127          & 13.665         & 22.597         & \textbf{27.65}              & \textbf{60.24}             & \textbf{82.36}             \\ \hline
\end{tabular}
\caption{\label{TRCDcomparison} TRCD results before and after pre-training on Arabic-SQuaD }
\end{table*}

\subsection{Discussion}
 The experimental results indicate that the
proposed pre-trained TunBERT model yields improvements, compared to
mBert, Gigabert and AraBERT models as shown in Tables \ref{TSACresults} and \ref{TECresults} for the sentiment analysis sub-task, Tables \ref{TADIcomparison} and \ref{TADcomparison} for dialect identification task, and Table \ref{TRCDcomparison} for the question-answering task.\\

Not surprisingly, GigaBERT as  customized BERT
for English-to-Arabic cross-lingual transfer is not effective for the tackled tasks and should be applied for tasks using code-switched data as suggested in \cite{lan2020gigabert}. \\

As AraBERT was trained on
news from different Arab
media, it shows good performances on the three tasks as the datasets contain some formal text (MSA).
The TunBERT was trained on a dataset including
web text, which is useful on casual text,
such as Tunisian dialect in Social media. For this reason, it performed better than AraBERT on all the performed tasks. 

We show that pretraining Tunisian model on highly variable dataset from social media leads to better
downstream performance compared to models trained on more uniform data. Moreover, results led to the conclusion that
a relatively small amount of  web-scraped dataset (67.2M) leads to downstream performances as good as models pretrained on a datasets  of  larger magnitude (24 GB for AraBERT and about 10.4B tokens  for GigaBERT).

This is confirmed with the QA-task experiments where the created dataset contains a small amount of dialect texts. The Arabic-SQuAD dataset was used to help with the missing embeddings of the MSA and to permit the finetuned model to effectively learn the QA-task by providing more examples of question-answering. The TunBERT model has overcome all the other models in term of exact match and recall.


\section{Conclusion}
In this paper, we reported our efforts to develop
a powerful Transformer-based language models
for Tunisian dialect: TunBERT. Our models are trained on 67.2 MB Common-Crawl-based dataset extracted from social media consisting of 500k sentences of text. When fine-tuned on the various labeled datasets, our TunBERT model achieves new SOTA on all the tasks on all datasets.
Compared to larger models such as
GigaBERT and AraBERT, our TunBERT model has better representation
of Tunisian dialect and yield better performances in
addition to being less computationally costly at inference
time.
Our models are publicly available for research\footnote{To preserve anonymity, a link to Github repository will be added to the camera-ready version if the paper is accepted.}. In
the future, we plan to evaluate our models on more
Arabic NLP tasks and further pre-train them to improve
their performance on the datasets where they
are currently outperformed.

On social media, Tunisian people tend to express themselves using an informal way called "TUNIZI" \cite{Chayma2020} that represents the Tunisian Arabic text  written using Latin characters and numerals rather than Arabic letters. For instance, the word "sou2el"\footnote{The word "Question" is the English translation.} is the Latin based characters of the word 
\textRL{\foreignlanguage{arabic}{سؤال}}. 
A natural future step would involve building a multi-script Tunisian dialect language model including Arabic script and Latin script based characters.

\bibliography{anthology,acl2020}
\bibliographystyle{acl_natbib}

\appendix

\end{document}